\documentclass[dvipdfm,a4paper,12pt]{article}
\begin{document}
\title{Comparison between the two definitions of AI\footnote{This work was supported by Bulgarian Ministry of Education, project DID02-28.}}
\author{Dimiter Dobrev\\
Institute of Mathematics and Informatics\\
Bulgarian Academy of Sciences\\
1113 Sofia, BULGARIA\\
e-mail: d@dobrev.com}
\renewcommand{\today}{March 22, 2013}
\maketitle

\begin{abstract}
Two different definitions of the Artificial Intelligence concept have been proposed in papers [1] and [2]. The first definition is informal. It says that any program that is cleverer than a human being, is acknowledged as Artificial Intelligence. The second definition is formal because it avoids reference to the concept of `human being'. The readers of papers [1] and [2] might be left with the impression that both definitions are equivalent and the definition in [2] is simply a formal version of that in [1]. This paper will compare both definitions of Artificial Intelligence and, hopefully, will bring a better understanding of the concept.
\end{abstract}

\section*{What is the basic idea?}

The idea behind the definitions of Artificial Intelligence in [1, 2] is as follows. If a program is intelligent it should manage well in an arbitrary world. This is a version of the popular wisdom that a clever person can handle any job. Certainly, the clever person will not be immediately successful in doing a novel job but after some training (learning). 

Therefore, before we evaluate how well a program does in a particular world, we should first allow for a certain time for training to pass. Only when the training is over, we could assess how well the program is doing.

We can make a comparison with humans. The first eighteen years of human life are considered as a period of training. If we do something wrong or even commit a crime, the punishment will be milder, because it is believed that we are still learning. The training period with animals is usually shorter; this is commonly associated with their shorter lives and worse living conditions. The rule that is often true about animals is `Learn fast or be eaten!'

How long is the training period for AI program and is it possible to say when exactly it is over?

Here comes the first difference between the definitions in [1] and [2]. The first definition assumes that the program's lifespan is infinite and therefore there is plenty of time for training. This is to say, if the training period has an arbitrary finite length, it is still infinitely shorter compared to the program's whole life, if the latter is infinite.

The approach in the second definition is different. It assumes that the program's lifespan is bounded and there is a parameter `maximum lifespan'. The program's success is evaluated on the basis of this finite lifespan. Another difference is with regards to the training period. It is not included in the second definition.The rule which is enforced is `Learn fast or be eaten'. 

The question arises, why the period of training is zero with the second definition.We set it to zero because we cannot say how long should this period be and when it will be completed. It is convenient to assume that there is no such period.

\section*{Reminding}

Let't to remind how were formulated the definitions of AI in [1] and [2].

Definition 1: AI will be such a program which in an arbitrary world will manage not worse than a human.

Here the worlds are not absolutly arbitrary because we suppose that fatal errors are not possible in these worlds.

Definition 2: AI will be such a program which IQ is bigger than 0.7.

Here the IQ is the averige succes acived by the program in some set of test worlds.

{\bf Nota bene:} When we call the Artificial Intelligence device a `program' we are not absolutely precise. A program is a piece of text, written on paper or another carrier. To start `living', the program must be run on a computer. There is a second inaccuracy, as well: while in Theory of Programs a program is usually associated with a computable function, here by program we will understand a transducer. A computable function takes as input some data and after some processing time returns output (if its program does not halt), while a transducer takes some information in and spits some out at every step. In [1] we stipulated that Artificial Intelligence is a step device. When we say here `the program' we mean this step device as a computer executing a program. In this paper we use the terms `program', `device' and `robot' synonymously when we speak about AI. However, for every step of the step device its program makes many small steps which correspond to the computer tacts. 

By `program's life' we mean all the steps a device has produced from the moment of switch on until we switch it off, or until infinity. Therefore, the program's life is a finite or infinite row of inputs and outputs for the device.

\section*{Abstract human being}

When in definition [1] we compare the intelligence of the program with the intelligence of a human being, we make the assumption that this is an abstract human being. A real human being could be sick, tired or bored, and that's why we will imagine an abstract human being who is always in good shape. 

What would happen if we compare the program with a real instead with an abstract human being? In such case, almost all programs would satisfy the definition, since even the most stupid program would do no worse than a human being. This is because all humans are mortal, while the program is immortal. At the beginning of his life the human will do better than the stupid program but this will continue as long as the human is still alive. From then on until infinity the program will do no worse than the dead human. If we calculate the average success of the human being and of the stupid program, they will turn out to be equal. That's so because, as we've already mentioned, any finite beginning of life is infinitely smaller than the entire lifespan.

Which program would we define as stupid? The most stupid program possible is the one who has an IQ of 1/2. (The definition of IQ is given below.) Programs with an IQ lower than 1/2 are not stupid, rather they are acting stupid or are following a losing strategy on purpose. Examples of programs with IQ of 1/2 are the arbitrary and the dead program. We call an arbitrary program one that makes moves randomly irregardless of its input, while a dead program is one that on every step performs one constant move irregardless of input.

If we want the human under comparison to be a real human being and not an abstract one, then we will have to assume that the program's lifespan is finite instead of infinite. To have enough time for training, we must assume that life is long enough but not that long for the human being to get tired, bored or to die.

\section*{Are the two definitions equivalent?}

Let's take a program that needs nearly infinite time for training (that is, its period of training is finite but in practice it is infinite). We would call that a retarded program. Examples for such programs are TD1 and TD2 in [2]. From the previous sections it follows that the definitions for AI in [1] and [2] are not equivalent because a retarded program is Artificial Intelligence according to the definition in [1], while it is not according to the definition in [2]. 

The retarded program does not match our idea of Artificial Intelligence. If our intention is to construct a robot to sweep the floor, it would not work for us if the robot needs a thousand years to learn how to sweep.

Let's check if there are other programs that formally satisfy the definition but do not match the idea of Artificial Intelligence. Such a program is the infinitely inefficient program, even tough it satisfies both definitions. Paper [2] describes such a program as Trivial Decision 5. This program would work if we had an infinitely fast computer, because in order to calculate a single big step it makes nearly infinite number of small steps (i.e. finite many but in practice infinite). Here, we call big steps the tacts of the step device and small steps are the computer tacts.

The problem is that by the definition of Artificial Intelligence we define a program without stipulating any requirements regarding its efficiency.

Both definitions for AI define a set of programs. These sets should coincide if the definitions were equivalent. The example of the retarded program shows that the definitions are not equivalent. 

There are two more reasons why both definitions are not equivalent:

Firstly, the definition in [1] is informal and dependent on people. That is to say, it does not define a particular set but rather a sort of fuzzy one. We said that we acknowledge as Artificial Intelligence those programs that are cleverer than humans. The reasonable question to ask with that statement is: `Who is the man we compare with?' A possible answer is that the program is cleverer than any human being, however, this would not define the set of these programs in a unique manner. For example, if we define the chess-playing program as a program playing chess better than the current top world champion, there will be still programs that will play chess better than top champion at one particular time and worse - at another. Therefore, a formal and an informal definition cannot be equivalent. 

Secondly, the definition in [2] is dependent on several parameters. This is to say, that it is not the case of a set of programs but rather a function that for different parameter values returns different sets of programs. 

The assumption made in [2] is that there exist parameter values for which the set defined by the definition is not empty and the programs within that set match our idea of Artificial Intelligence. 

Certainly, not all programs within that set match our idea, however. We have already mentioned the problem with the infinitely inefficient program. Another problem that appears with the definition in [2] is the `cramming' program. The problem is that in [2] it is assumed that the worlds we are interested in are finite in number. Therefore, a program can be created specially for these worlds. We can illustrate this problem with students taking their exams -- those who have memorised by heart all possible exam topics will pass the exam but they won't be able to solve any problem outside of that range. 

The assumption is that if we select the shortest and the most efficient program out of those that satisfy the definition, it will match our idea of Artificial Intelligence. It is necessary to limit the programs in both length and efficiency because the infinitely inefficient program is rather short, whereas the `cramming' programs are rather efficient. The `cramming' program is shorter than the Artificial Intelligence one for a small number of worlds, while the Artificial Intelligence program is a shorter than the `cramming' one, if the number of worlds is sufficiently great.

\section*{Lifespan}

The program's lifespan is the primary parameter, on which the second definition depends. Once we give up infinite life, then we should limit it and set a parameter indicating the life expectancy. To make it simple, the program's lifespan in [2] has been fixed to 100 games, each one no longer than 1000 steps. 

Giving up infinite life, we let go of the retarded program. Another advantage is that it is no longer required that the worlds do not allow fatal errors. This requirement was important in [1] because in order to ensure there is sufficient time for training in a program's life, we had to exclude the possibility of making a fatal error that would ruin the program's possibility for success for its entire lifetime. In that case the program could safely make mistakes as each one wouldn't be fatal and could be overcome. 

{\bf Definition:} Fatal error is a group of internal states of the world, such that when we enter this group we can no longer exit it. If there is an exit, the error would not be fatal. Besides, the world in this group should be worse than the world outside of the group (i.e. the rewards received there should be relatively lower). If the world in the group is not worse, then the fact that we happen to be there would not be a mistake. 

Another possible definition for fatal error is the following. If at each moment of life we calculate the maximum anticipated success for that life trajectory (i.e. the result of the Success function), provided that from that moment onwards we play (we live) by the best possible strategy, then fatal error will be referred to as a step after which this number is decreasing. 
 
Having assumed that life is finite, it is not needed to assume that there are no fatal errors in the world, because our time for training is anyway limited. When life is finite, a common mistake can be equal to fatal, because time may not be sufficient to fix it. 

Thus, it is natural to assume that life is finite and we are looking for a program that can manage well within concrete lifespan rather than looking for a program suitable for arbitrary lifespan. On the other hand, it is inconvenient if there are parameters included in our definition. It would be better to define Artificial Intelligence as a program independent of anything. That is to say, a sole program, irrespective of the anticipated lifespan. 

However, life expectancy is an important parameter that influences actor's strategies. Let's consider human behavior in war time, during natural disasters or other catastrophes. When life expectancy becomes shorter, human behavior changes significantly. This is expressed mainly in the tendency to take greater risks. You may also notice that young people are braver than older ones. A possible explanation of this observation is that the young are more willing to experiment and take risks while the adults prefer stability and security because they estimate that there is not so much time anymore for experimenting. Thus, we can say that life expectancy definitely influences behavior of people and their life strategies.

\section*{Arbitrary world} 

The first definition requires for the Artificial Intelligence to manage as good as a human being in an arbitrary world. This requirement is so strong that it seems that there is no program that can meet the requirement. The set of programs that satisfy the definition may prove to be empty. 

Let's try to construct a world which is too complex for any program to succeed in it but not for the human. Imagine a world where robots are not liked. In such world, if you are recognized as a robot you immediately score low. However, if you are considered human you score high. It seems that this is the world where humans will do better than robots. Let's remember the definition for a world; there are two functions (World and View) defining the world. They are absolutely arbitrary functions and we can presume that they return success when there is a human being living in this world, respectively, they return no success in the case of a robot. Still, the world is not God and there is no way to know if its inhabitant is human or robot. The world will know who is who based on the acts of the actor. That is, if the robot behaves like a human and acts appropriately then the world will be deceived and will accept it as a human. In this case, the program is forced to play a game of imitation. Such a game was proposed by Turing as a test for Intelligence. It is to be concluded that if a program meets the definition in [1], then it is satisfactory to the Turing test -- maybe not immediately but after it has taken some time for training. 

{\bf Question:} Is it possible that the world recognizes the robot (while it is still on training and has not started to act like a man) and starts scoring low from that first moment of recognition till infinity? The answer is: No, because only worlds without fatal errors are considered and this world does not meet this requirement. 

Does that mean that the definition in [1] is equivalent to the Turing test? Not, if we train the program in [1] to pretend being a human, then it will satisfy the Turing test but only after training. Is it possible that the program satisfying the Turing test be trained to do well in an arbitrary world? The answer is: most likely not. If the program can pretend to be human, then it can be trained for any world. However, it should rather pretend to be stupid and hide its intelligence or otherwise it will betray the fact that it is a robot and not a human. If it is forbidden for the examiner to punish excessive intelligence, then the definition in [1] and the Turing test will be equivalent.

\section*{Impossible World} 

Is it possible that a world is so complex that there is no program that could understand it? Yes, it is. For example, let the world generate an infinite row of zeros and ones. The Artificial Intelligence program has been given a task to make a guess what comes next (zero or one). Let the function describing this infinite row is not computable. Then, there is no way for the program to calculate and say with confidence which number will follow. This is true about the human, as well. Nevertheless, the program and the human will find different dependencies. One could be that the zeros are more than the ones, or another -- that one is more likely to come after a zero, etc. 

It is not necessary that the Artificial Intelligence should understand the world at 100 \%. What is important is to understand the world better than the human.

\section*{IQ (Intelligence Quotient)}

The first definition compares the intelligence of the program with that of a human. The second definition cannot make this comparison, because we want the definition to be formal. Because of that, it is necessary to introduce an independent assessment of IQ by which we could define Artificial Intelligence. We will say that we acknowledge as AI those programs whose IQ is above certain value. This value was taken to be 0.7 in [2] but this choice has been largely arbitrary. It is rather correct to say that certain IQ exists and the programs more intelligent than this level are acknowledged as Artificial Intelligence. 

We introduce the function Success which returns a number in the interval [0, 1] for each particular life. This number makes assessment of the program's success in the particular life. Afterwards, the IQ is calculated selecting a set of test worlds, running the program to live a life in each one of these worlds and calculating the average success of the program in all its test lives. 

Thus, the IQ is the average value of Success function calculated based on the set of test worlds.

\section*{World Complexity}

Another substantial difference between both definitions is that the first considers all possible worlds; whereas the second limits the sets of the worlds to a finite number of test worlds (the assumption is that the test world is computable with fixed level of complexity). This fixed level of complexity is the second parameter in the definition.

How did we select the set of test worlds?

Something similar has been done in paper [1]. There it was proposed that we prepare a test consisting of finite or countable number of test worlds. The idea is to acknowledge as Artificial Intelligence the program that can manage in all these worlds. Paper [1] has proposed that these worlds be prepared by a human but we want to be as formal as possible in paper [2] and therefore we will define the set of test worlds in a way that does not rely on human actions. The other difference is that in [1] we want the program to pass all exams, i.e. to manage in all test worlds, whereas in [2] we want the average success (i.e. IQ) to be greater than 0.7. Why we want to have less in [2] than in [1]? Because, if the problems are prepared by hand we would want the program to solve them all, but if the problems are randomly generated then some of them will be insolvable and therefore we cannot require from the program to solve all problems. 

What is the set of test worlds that we could use for the calculation of the IQ of an arbitrary program?

The first natural possibility is to take the set of all worlds. This set is infinite, even uncountable and seems too large (it is not clear what should be the weight of different worlds in the set). The first thing we see is that many of the worlds are indistinguishable (i.e. their tree of the world is the same). That's why we resort to the next idea: to take the quotient set, i.e. the set of all possible trees of the world, and make it our set of test worlds. This set is again uncountable, but considering the fact that we limited the lifespan, we see that the set of these trees is even finite (more precisely, the set of the trees of determined worlds is finite. It is finite with the undetermined worlds, as well, because the branches are equally probable to happen -- see the definition of TM\_W in [2]).
 
This set is not suitable (although it is finite) because in such a case anything is possible! How would the world evolve for the next step? In fact, it could evolve in any way. Anything is possible, indeed, but far from anything is probable. If we accept this set as the test one, then any continuation will be equally probable and the past will be of no significance. This totally contradicts our idea of Artificial Intelligence which says that the device gathers experience and undergoes training. Thus, what has happened in the past is important. 

This is the point to apply the principle known as `Occam's razor' stating that the simpler model is more probable than the complex one. Therefore, the simpler world is more probable than the complex one. If we are to discuss the complexity of a world, then we would introduce a description of the world and define the complexity of the world as the length of the shortest possible description. 

We have used Turing machines to describe the worlds in [2]. This is not the most suitable model in case you try to make a real program which satisfies the definition. However, here it works as a theoretical model of computability. Still, we do not want to be limited within the set of the determined worlds and this is why we have introduced undetermined Turing machines. Thus, our test worlds are the computable worlds generated by the undetermined Turing machines. 

We suggest next the set of test worlds to be taken as the set of the undetermined Turing machines, inclusive of all such machines regardless of their size. Could we choose a particular size and do only with the machines of that size? The answer is: rather yes. 

If we take all Turing machines we should give them different weights since there is no way that an infinite number of machines are of equal weight and the sum of their weights equals one. Having decided what will be the weights of the different machines, there are two options to go for: either the average size (length) of the machines is a particular number, or the average size is infinity (depending on the weights we have finally chosen). If the average size is finite, then we can assume that instead of using all Turing machines as test ones, we would use only those whose size is the average. This is not the same but is almost the same. If the average size is infinity, then there exist an $N$ such that all machines with greater size would have almost no influence on the average. Therefore, is we chose an $\varepsilon$ that seems to us small enough to be ignored, then such $N$ exists that the machines longer than $N$ will influence less than $\varepsilon$ of the average success. Then we can decide that $N$ is the size of the test machines and the result will be similar to what we would have if we consider all machines with their respective weights.

Next question: If we have decided on a particular $N$, should our test machines be these of size smaller or equal to $N$ or those with the size of $N$ exactly. The answer is: there is no need to include the shorter machines, because each machine with size $N-1$ has many equivalent machines with size $N$ (because we can add a state that is not necessary).
 
So far, so good. We have decided that the test worlds will be the computable worlds that are computed by an undetermined Turing machine having the size of $N$. This is the next parameter that our definition depends on. In [2], we decided on a particular value of 20 for this parameter. We choose that all test machines will participate with equal weights (this is possible because the set is finite). 

Does the set chosen in this way correspond to the principle of Occam? Are the simpler worlds more probable than the complex ones? The answer is yes. Indeed, all machines participate with equal weights but the simple machines come with a large number of equivalent machines (which compute the same world), whereas there is not a single equivalent machine for the most complex ones (certainly, not of the same complexity, in this case with complexity value of 20). This is to say that the simpler the world is, the more machines with size 20 happen to compute it and the more this world would influence the average value of Success function. We refer to this average value as IQ.

\section*{Which is the suitable model?}

We have already said that Turing machines are not the suitable model to describe the world. We would like to have simple dependences in the world that are on the surface and easy to be discovered; whereas the deeper we go, the more and more complex dependences become. The Turing machine is a dependence that may appear to be rather complex but once you have understood it, you have understood the world. The case of the undetermined machines is more suitable because the randomness is an infinitely complex dependence. Therefore, we will never understand this dependence because once we understand it, it will become pseudorandomness (take the example of pseudorandom numbers generated by a computer). 

Is it possible that a complex Turing machine is partially described by means of simpler dependences? This is possible but it is not typical for the Turing machine model. In this model usually, you either understand the whole world or you don't understand anything. 

If we are looking for a world model of the type of à determined machine, then very soon (i.e. after very short life experience) it would turn out that the first Turing machine corresponding to that life experience is so complex that it is virtually impossible to be found. The advantage of the undetermined machines is that we will always be able to find such world model (no matter how long is the life experience). It is a different question how adequate is this model and how good it will work for us, because the undetermined machine does not say what will happen on the next step, but it rather says that this or that can happen. In the best case, it provides the probability of having this or that happening. 

Well, which is the suitable model of the world? We should think of the world as a union of different factors that may be connected but are largely independent. Certainly, we will need a better model if we want to create a particular program satisfying the definition of Artificial Intelligence. However, this is not important for the definition itself.

\section*{Work of other researchers}

The occasion to write this paper is publication [5] where two swiss scientists attempt to generalize the definitions in [1, 2]. Their idea is to get rid of the parameters, which the definition in [2] depends on and to get to a new concept of IQ independent of any parameters. 

They remove the restriction on lifespan and assume that life is infinite. In their discussion the initial part of the life trajectory is the most significant. They assume that the rewards become lighter at each next step, having been multiplied by a discount coefficient

Surely, the coefficient of discount is also a parameter, and they have simply replaced one parameter with another. What is more, their choice contradicts the idea that the beginning of life is not important. What is important is what happens after the program has been already trained. Their presentation assumes the beginning of life as the most important part. After some time the life trajectory has suffered so much discounting that in practice, it is not important what the program is doing anymore. Certainly, in [2] there is a moment (the maximum length of life) after which it is not anymore important what the program is doing, but at least until that moment all rewards are of equal importance. This is to say, in our case `it runs, it runs and stops', whereas in theirs `it fades, it fades, it fades and like this to infinity'. 

We have to acknowledge that the authors of [5] have understood that the coefficient of discount is a parameter which their definition depends on and have, therefore, proposed a second version. It would have been better if the second version have not been proposed, at all, as it has resulted in meaningless outcome compromising the whole paper. More, about the second version, has been written further below (see Striking mistakes 3 and 4). 

The other parameter that the authors of [5] have tried to get rid of is the world complexity (the number of states of the Turing machine which generates the world). We have decided on a particular complexity. They have preferred to sum up all complexities having used a coefficient of discount 1/2. This results in having average complexity of 2 in their case (this parameter has been picked to be 20 in our case). This is to say, that if they want to allow higher values of average complexity, they will have to replace the number 1/2 with a different parameter. Therefore, they replace one parameter with another, again. 

They get rid of the parameter 0.7 by avoiding to say what Artificial Intelligence is. They define IQ but they do not say how big the quotient should be in order for a program to be acknowledged as Artificial Intelligence. 

Unfortunately, our Swiss colleagues have failed to quote the Bulgarian primary source. Another problem is the fact that they have not managed to understand many details of the original papers and, therefore, there are many mistakes and inaccuracies in the resultant text.

\section*{Striking mistakes}

Here are six of the most striking mistakes made in [5].

1. The authors state that the world is computable according to the thesis of Church. Indeed, [1] states that it follows from the thesis of Church that the Artificial Intelligence is a program but not that the world is a program, too. Is the world computable or not, is it determined or not -- these are questions whose answers we do not know and will never know. This is something that cannot be verified because there is no experiment that can result in an answer to any of these questions. (The question whether the world is determined has been considered in details in [3]. The question whether the world is computable is analogous). 

2. The authors have defined IQ in such a way that it is infinity for each program. They have not considered the fact that the number of programs grows exponentially with the increase of their length. This can be regarded as a mistake by oversight, moreover that it is clear how it can be fixed. (In their case they sum up the programs' success achieved in different lives, i.e. the values of function Success. This sum does not require that each addend is multiplied by 1/2 raised to the power of the complexity degree. Rather, one should take the average for the respective complexity and multiply it by the same discount). We said above that the average complexity of the world is 2, having assumed that this mistake is fixed. If [5] remains unchanged, then the average complexity is infinity and the sum that we refer to as IQ is infinity, too. Thus, if this mistake is not fixed, the IQ concept becomes meaningless. 

3. The most serious problem in [5] is the second version of definition that was proposed in order to avoid the coefficient of discount. The set of worlds differs in this version and respectively the Success function differs too (the Success function returns assessment of the success that has been achieved in each life trajectory). 

Let's summaries the outcome in [1], [2] and both versions in [5].

In [1] is considered an infinite life in a set of worlds without fatal errors. In [2] is assumed that life is finite. The first version proposed in [5] describes life as infinite but fading which is the same, as if the life is finite. The second version proposed in [5] assumes infinite life in a world where all errors are fatal. The problem in this case is not that there could be a fatal error (fatal errors do not interfere in [2] and will not interfere in this case, as well). The problem is that all errors are fatal. This is to say, that there are no fixable errors. Humans do not learn from their fatal errors. Perhaps, they have learned from the fatal errors of others but not from their own. That's why this concept contradicts the idea of learning. 

The Success function is monotonically increasing in the second version proposed in [5]. It was defined to be the sum of all rewards which are numbers in the interval [0, 1]. It is natural to consider that the Success function can be increasing and decreasing during a lifetime. Life steps at which it starts to decrease can be considered as a fixable mistakes. The authors of [5] have preferred to have this function monotonically increasing which means that the device cannot make fixable errors. The only possible mistake is of the type `lost profits' and this mistake is always fatal because once the profits are lost there is no way to bring them back. Besides, there is no feedback in the `lost profits' case. When the Success function changes, the program will know that, but there is no way to know when it is missing on profits. Eventually, it might get to know that in the future, but since life is infinite there is really no way for the device to know -- it will always hope that the profits are not lost and will soon materialise.

4. The decision of the authors of [5] to limit the sum of the rewards is rather strange and illogical. It reminds me of one maths professor during my undergraduate years. All of us students at the time believed that he had a limit on the A grades. We thought that one should be among the first to be examined because he will run out of A grades and no matter how much knowledge you demonstrate, if you are among the last to be examined, you will not get an A. 
 
This limitation is imposed, in their case, so that the Success function will be in the interval [0, 1]. Instead of distorting the world in such a horrible way, it would be better if the Success function is the arithmetical mean of the rewards (as it was done in [2]) instead of being the sum of the rewards. This would place the Success function in the interval [0, 1], and it would lift the requirement to be monotonically increasing. 

The conclusion is that in their second version of the IQ definition the authors of [5] use worlds where training is impossible. The success of the device in such a world depends solely on its luck, however if the worlds are sufficiently many, luck ceases to have an effect. Thus, all programs will be equally intelligent. 

The question to ask is whether the authors of [5] have managed to understand that both [1] and [2] consider a device that is being trained and will achieve good success as a result of the training, or they rather believe that it is born trained. As a matter of fact, the beginning of [5] says that the device should be given sufficient time for training, however, later they propose two versions of the definition which contradict this idea (especially the second version). 

It is true that with the definition of Turing, there is a device that was born trained, but this concerns a particular world. The device can be born trained for a particular world but there is no way that it was born trained for every world. 

5. We can consider as a mistake the fact that the definition of a world in [5] has been altered. The world in [1] has got a set of internal states and a function indicating how to transfer from one state to another. As you know, there is a tree of the world corresponding to each world. It was the tree of the world that has been regarded for the definition of the world in [5]. This is the same as if we do not consider functions in mathematics but only graphs of functions. We can somewhat justify the authors of [5] because they make their attempt to improve somehow the definition of AI, however their change implies that they have not understood our main idea. We suppose that the world has got some structure and the device attempts to understand that structure. They deny the structure of the world and this is a mistake.
 
They make a similar mistake when they define the device, as well. The device is a program for us, whereas for them it is a strategy. Certainly, there is a strategy corresponding to each program (but not vice versa). Still, to consider the device as a strategy is a mistake, as thus we presume that it has not got internal states, i.e. no memory. This mistake is very common among researchers working in the area of AI. Many of them look for the AI in the set of the functions, meaning that for them AI is a device without memory. The strategy is also a function, whose input argument is the entire life experience (life history). At first sight, it looks as if we do not need a memory as if we have as input the entire life experience, but this is not true. 

Imagine that at some point (based of your whole life experience) you decide to go to the fridge and grab a beer. Ten seconds later, you see that you are on your way to the fridge but you do not know if you are going to get a beer or milk. It is true that you can rely on your whole life experience but this will not answer the question. If you were told to fetch a beer, this can be extracted from your life experience, but if you have, yourself, decided to grab a beer, you would not remember it (because you have no memory) and there is no way that you extract it out of the life experience. This is to say, the memory is needed. We have to note that it is absurd that the device will make a decision at any step based on its whole life experience (because this is a huge amount of information). It is more reasonable to assume that it decides based on its internal state and the immediate input received in the last step. 

6. The last comment to make with reference to [5] is the strange argument about the question whether the rewarding is to be part of the world or part of the device. The rewarding has been confidently treated as part of the world at the beginning of [5] and the dilemma about its belonging at the end of the paper is rather odd. The authors answer this question themselves saying that if the students are allowed to mark their knowledge themselves they will all have excellent results. 

Nevertheless, the confusion of Shane Legg and Marcus Hutter about the belonging of the rewards is reasonable. They consider the human being and ask themselves aren't the rewards in that case the feelings of pain and pleasure (with food, sex, music and other sources of pleasure). The answer we would give is that the human has not built-in rewards. Success with regards to the human being is evaluated by the world through the process of evolution. Evolution defines success in our world for those who manage to survive in it and to reproduce. A human does not receive his or her rewards system readily. Actually, humans never receive it. If one has lived in compliance with the evolution's principles, about which he is not aware, then he will survive long enough to be inherited in the next generation. This is the reason why people are looking for the meaning of life all their lives (they are looking for it, because they do not know it). As long as the feeling of pain and pleasure are concerned, this is not the meaning of life but an instinct. Thus, humans are born with some knowledge. They instinctively know that pain is bad, whereas pleasure is good. These instincts should not be trusted blindly. For example, the feeling of pain when the dentist pulls a tooth is a misleading signal (as far as the dentist pulls the right tooth, of course). The bitter taste in food is instinctively perceived as bad but with time people get to like the taste of coffee and of beer, for example. The feeling of pleasure is also often a misleading signal. 

Despite all the remarks we have made, the papers of Shane Legg and Marcus Hutter are very valuable to us, because they are the first to acknowledge the definitions in [1, 2]. Furthermore, the analysis of the mistakes made by our Swiss colleagues is also useful because it indicates what has not been explained sufficiently well. It is obvious that Shane Legg and Marcus Hutter have done serious work on this subject and it was useful for us to study their attempt to improve our definition. Where they have not understood the definitions of [1, 2], this is our fault, as we seem to have poorly explained them before. The present paper was prepared on the basis of this analysis and we hope that it will be helpful for all researchers working in the field of Artificial Intelligence.

\section*{Getting rid of the parameters}

Could we, after all, get rid of the parameters in definition [2]? We could get rid of the parameters taking part in the definition of IQ, but I cannot think of a way to get rid of the coefficient that tells us which is the minimum IQ sufficient enough to acknowledge the program as AI. 

To remove the parameters in the definition of IQ, let's look at the IQ as a function of the lifespan and the world's complexity. We could take the limit of that function in the case when these two variables approach infinity, and take this limit to be the new, parameter-free, IQ. 

{\bf Note:} Actually, [2] mentions yet another parameter. This is the maximum number of small steps that the Turing machine of the respective world is allowed to make in a single big step. We could assume that the complexity of the world depends on two parameters -- the number of states of the Turing machine and the maximum number of small steps for a big one. That means that the IQ definition depends on three rather than on two parameters.

If we define IQ through a limit to infinity, the question is whether such limit of this function actually exists. Our first concern is whether that limit would not turn out to be infinity. Let me remind you that the Success function was calculated as the arithmetical mean of the rewards gained for the victory, loss or draw, which are respectively 1, 0 and 1/2. This means that the function is in the [0, 1] interval. It follows that IQ is also in the [0, 1] interval because this is the mean value of the Success function. Therefore, the limit of IQ in the case when its arguments approach infinity is also in the [0, 1] interval. 

Our second concern is whether that function would not turn out to be divergent, i.e. its lower limit to turn out to be strictly smaller that the upper one. Is it possible for a world to exist in which success jumps up and down again and again in the course of life and never converges to any specific value? Imagine a world in which with the first step you gain victory, with the next two steps -- loss, with the next 4 steps victory again, and with the next 8 -- loss again.

Worlds like these are, of course, very few and cannot affect the IQ value, but is it possible for a program to exist, whose IQ jumps up and down again and again depending on the lifespan? Imagine a program which is quite clever and knows what's going on, but nonetheless now plays to win and now plays to lose. We will again use the sequence of intervals 2n and will assume that in each following interval the program changes its behavior. 

Is it possible for a program to have a good IQ when the complexity of the world is low, but to show a rapid drop in its IQ when the complexity of the world increases? Yes, this is the so-called `cramming' program. This program does well in a small number of worlds, provided it has been designed specifically for the worlds in question. However, when the worlds grow in number and in complexity, the `cramming' program fails. 

Therefore, it is possible that the IQ function is not convergent. In this case the new IQ would be chosen as the arithmetical mean of the IQ function's lower and upper limit.

If we accept this new definition of IQ, we will loose the best feature of this function -- the fact that it is computable. Nonetheless, we do not compute the exact value of this function due to the combinatorial explosion caused by the large number of possible worlds and possible life trajectories. Instead, we calculate its approximate value by using the statistical method: picking randomly a small part of all test worlds. This is our statistical sample within which we let the program live one test life in each world.

We can approximately calculate the new IQ function as well, by taking at random one countable sample of worlds and putting them in line. We will want that the sequence of test worlds to show growing complexity. We will let the program live one test life in each sample world, making lives longer at each run. For every test live we will calculate the Success function. If we take the average Success from the first N lives then we will have a sequence of interim results, which will converge to the limit of IQ (i.e. the new IQ, if the old IQ has a limit). If the function IQ has no limit, then the sequence will not be convergent but its upper and lower limits will coincide with the lower and upper limit of IQ.

Of course, we cannot calculate the limit of this sequence. We have to put up with approximating calculation. We can choose an N which is big enough and suppose that the limit of the sequence is approximately equal to its N-th element. In fact, this is the same as if we pick a big enough complexity of the world and big enough lifespan and calculate for them the value of IQ of the program.

The question is whether to define the new IQ as the limit of IQ in the case when the complexity of the world and the lifespan approach infinity, or whether it is better to assume that we've selected big enough values for those parameters, and to take this as our new IQ. [2] adopts the second solution.

Concerning the coefficient that determines the minimum IQ sufficient enough to acknowledge the program as AI, this value must be somewhere between 1/2 (the intelligence of the most stupid program) and the intelligence of the smartest one, i.e. the intelligence of Trivial Decision 5. It is advisable to take the IQ of a human being, since for us a program with a sufficient intelligence is the program smarter than a human being. The problem is that we do not know the value of human IQ. We could try calculating its approximate value by making tests with human beings. I have some experience experimenting with my students, and the result is that students cannot manage even with a simple world such as the Tic-Tac-Toe game. It is true that with this case there was some over-coding, which made the world too complex and prevented the students from understanding it. It is also true that training is supposed to continue your entire life, while my students tried to understand this world in only 45 minutes. I assume that if the students have passed training in solving problems of the kind, they would show much better results. However, the results from my experiment with students cannot be accepted as reliable and that, in general, human IQ is difficult to measure following experimental methods.

Of course, measuring human IQ is completely useless. What is important for us is to know what AI is, and these are the programs of sufficient intelligence. The more intelligent a program is, the better.

\section*{Who is more perfect?}

The definition of IQ creates an order: who is more intelligent and who -- more stupid. When we change the definition of IQ, we in effect alter the order that this definition gives. Thus a program could be ``smarter'' than another under one definition of IQ, and not so under the other definition. If we choose the IQ to be the limit in the case when the lifespan and the complexity of the world approach infinity, then we would prioritise for the programs learning at a slower rate rather than the programs learning at a faster rate. 

One of my father's classmates [7] had a theory that the slower one creature develops, the more perfect it is. He gave examples with animals and humans, claiming that humans are more perfect than animals. Yet another of his arguments was that most renowned people did not show much brightness when they were children. He also compared man and women, saying that girls develop faster than boys but men achieve better results in the field of abstract thinking.

I agree with his observations, but I cannot agree with his conclusions. First of all, I do not agree that humans are more perfect than animals. If we compare the physical strength, fastness, endurance to cold, famine and pain, we will see that humans significantly lack in those qualities compared to animals. We humans are also inferior to animals in regards to beauty. It is true that we like women, but that's because we are instinctively programmed to like them, and not due to some objective reasons. Women do not have colorful feathers or something else that would impress an animal of another species. Take for example the udder of the cow -- for the bull it would probably be a source of inspiration for a poem, but for us humans it has no beauty at all. 

This article is dedicated to the intellect and therefore we are to compare humans and animals in terms of their intellect, not taking into account their other features. Can we claim that humans are cleverer than animals? Comparing two brains is like comparing two cars. If one is faster and the other more powerful, which one is the better? The brain has so many characteristics that it is difficult to decide which brain is better. For example, the long-term memory of elephants is much better than that of humans. Experiments show that some monkeys have much better short-term memory than humans. If we assume that the most important characteristic of the brain is the ability of abstract thinking, then it might turn out that octopuses are much more capable than us humans. Of course, octopuses are very stupid animals but this is because abstract thinking is not all. Octopuses have no relations between generations and they are self-taught. To compare a human being with an octopus, a human must have grown without parents and educators. 

Shortly, I do not agree that people are more perfect than animals. I don't agree either that abstract thinking, which is better developed with men, is more important than intuitive thinking -- more developed with women. It is true that with abstract thinking, when we have a solution, we know how we've come to that solution, while intuitive thinking gives us a solution but we do not know how we've reached that solution. That's why, often when we talk about reasoning, we mean abstract thinking, but there are areas where intuitive thinking gives a much better result than abstract one. In areas like these women are much more successful than men.

As far as the so-called renowned people are concerned, I also cannot agree that they are more perfect than the rest of us. It is no coincidence that most geniuses have not been recognized during their lifetime. Can you imagine how many geniuses there are who are not recognized even after their death? Geniuses that all their life have been considered idiots and are long forgotten. 

It follows that it is not a good idea to change the definition of IQ and take the program that is most suitable for the case when the lifespan approaches infinity. Thus, the retarded program will become one of the cleverest ones, and as we've already mentioned, this program does not satisfy our idea of Artificial Intelligence. It's better to stick to IQ defined as the most suitable program for the specific lifespan.

Different tasks take different tools. Our idea of finding a universal tool that is good for any task is praiseworthy but very often -- difficult to attain. Very often one quality is at the expense of another. With a real brain, when one part is better developed, this is at the expense of the rest (and that's because the real brain must fit into the real skull). When we give a description of the artificial brain, we can assume that we have no physical limitations and that all parts could be developed to a maximum extent; however, the training time is a limited resource with both the real and the artificial brain.

 ``Time'' in regards to the artificial brain means the number of steps, while for the real it means seconds, hours and years. Let's assume that the real brain makes 24 steps per second (as with the frames in cinema); thus, the idea of time with the real and the artificial brain will be one and the same.

\section*{Acknowledgements}

I want to thank Professor Dimiter Skordev and Professor Tinko Tinchev for their comments and advice regarding this paper. Furthermore, I would like to thank Professor Skordev that in 2004 he challenged me to compose definition [2]. He said the following regarding definition [1]: `This definition is not formal, at all. You claim that it can be formalized but what is not clear is how it can be done!' This was a constructive criticism and its result was definition [2]. 


\section*{References}


.

[1] Dobrev D. (2000) {\it AI -- What is this}, In: PC Magazine -- Bulgaria, November'2000, pp.12-13 (www.dobrev.com/AI/definition.html).

[2] Dobrev D. (2005) {\it Formal Definition of Artificial Intelligence}, In: International Journal ``Information Theories \& Applications'', vol.12, Number 3, 2005, pp.277-285 (www.dobrev.com/AI/).

[3] Dobrev D. (2001) {\it AI -- How does it cope in an arbitrary world}, In: PC Magazine -- Bulgaria, February'2001, pp.12-13 (www.dobrev.com/AI/).

[4] Legg S. and Hutter M. (2005) {\it A Universal Measure of Intelligence for Artificial Agents}, In: Proc. 21st International Joint Conf. on Artificial Intelligence (IJCAI-2005), pages 1509-1510, Edinburgh, 2005. 

[5] Legg S. and Hutter M. (2006) {\it A formal measure of machine intelligence}, In Proc. 15th Annual Machine Learning Conference of Belgium and The Netherlands (Benelearn'06), pages 73-80, Ghent, 2006.

[6] Legg S. and Hutter M. (2007) {\it Universal Intelligence: A Definition of Machine Intelligence}, Minds \& Machines, 17:4 (2007) pages 391-444.

[7] Nikola Konstantinov Bachvarov (1932-1977)


\end{document}